%% file: main.tex
\pdfoutput=1

\documentclass[11pt]{article}
\usepackage{booktabs}
\usepackage{graphicx} 
\usepackage{algpseudocode}
\usepackage{acl}

\usepackage{times}
\usepackage{latexsym}

\usepackage{multirow}
\usepackage[inline]{enumitem}
\usepackage{enumerate}  
\usepackage{amsfonts}

\usepackage{xspace}
\usepackage{amsmath}

\usepackage[T1]{fontenc}

\usepackage{amssymb}
\usepackage{newfloat}
\usepackage{listings}

\usepackage[utf8]{inputenc}

\usepackage[dvipsnames]{colortbl}

\usepackage{xcolor}
\usepackage[normalem]{ulem}

\usepackage{babel} 
\usepackage[toc,page]{appendix}
\usepackage{makecell}
\usepackage{soul}
\usepackage{verbatim}
\usepackage{arydshln}

\usepackage{adjustbox}
\usepackage{footmisc}
\usepackage{microtype}
\usepackage{bbm}
\usepackage{wasysym}
\usepackage{tikz}
\usepackage{tcolorbox}
\usepackage{siunitx}
\usepackage{pifont}
\usepackage[fixed]{fontawesome5}
\usepackage{hyperref}

\usepackage{inconsolata}

\newcommand{\headernodot}[1]{\vspace*{1mm}\noindent\textbf{#1}}
\newcommand{\header}[1]{\headernodot{#1.}}

\usepackage[nameinlink]{cleveref}
\newcommand{\instrctionsize}{\footnotesize}
\newcommand{\model}{PSI\xspace}
\newcommand{\fullmodel}{\textbf{P}rinciple-based \textbf{S}elf-\textbf{I}nstruct\xspace}
\newcommand{\first}{Multi-level Principle Generation\xspace}
\newcommand{\second}{Principle-based Instance Generation\xspace}

\newcommand{\eg}{\emph{e.g.,}\xspace}

\newcommand{\ignore}[1]{}

\newcommand{\dubbelneer}{$^{\blacktriangledown}$}

\newcommand{\cbkgrnd}{\cellcolor{blue!15}}
\title{Evolution without Large Models: Training Language Model \\ with Task Principles}

\author{
 \textbf{Minghang Zhu\textsuperscript{1}}\thanks{Equal contributions}\space\space\space
 \textbf{Shen Gao\textsuperscript{2}}\footnotemark[1]\space\space\space
 \textbf{Zhengliang Shi\textsuperscript{1}}\space\space\space
 \textbf{Jiabao Fang\textsuperscript{1}}\space\space\space
\\
 \textbf{Pengjie Ren\textsuperscript{1}}\space\space\space
 \textbf{Zhaochun Ren\textsuperscript{3}}\space\space\space
 \textbf{Zhumin Chen\textsuperscript{1}}\thanks {Corresponding author}\space\space\space
 \textbf{Shuo Shang \textsuperscript{2}}
\\
 \textsuperscript{1}Shandong University, Qingdao, China \\
 \textsuperscript{2}University of Electronic Science and Technology of China, Chengdu, China \\
 \textsuperscript{3}Leiden University, Leiden, The Netherlands
\\
\small{  \texttt{ \{mhzhu,shizhl,jiabaofang\}@mail.sdu.edu.cn, shengao@pku.edu.cn   } } \\
           \small{ \tt{ z.ren@liacs.leidenuniv.nl, jedi.shang@gmail.com, \{chenzhumin,renpengjie\}@sdu.edu.cn } }
}

\begin{document}
\maketitle
\begin{abstract}
A common training approach for language models involves using a large-scale language model to expand a human-provided dataset,  
which is subsequently used for model training.
This method significantly reduces training costs by eliminating the need for extensive human data annotation. 
However, it still faces challenges such as high carbon emissions during data augmentation and the risk of data leakage when we use closed-source LLMs.

To address these issues, we propose \model, a self-evolution method for language models fine-tuning.
First, we introduce the \first, which enables a large-scale model to summarize task-completion principles based on a small amount of task data. 
Then, we propose the \second, in which a smaller-scale language model uses these task principles to generate a large amount of data. 
This data is then used for model training. 
Experimental results on several benchmarks show that our proposed method significantly improves model performance compared to directly using a smaller-scale language model to generate data. Additionally, since we only use the large-scale language model to generate the task-completion principles, the carbon emissions associated with training the model are greatly reduced.
\footnote{\href{https://github.com/ZMingHang/PSI_method}{\faGithub~GitHub}}.
\end{abstract}

\input{sections/01-Introduction}

\input{sections/02-related-work}
\input{sections/03-method}

\input{sections/04-experiment}
\input{sections/05-results}

\input{sections/07-conclusion}

\input{sections/limitations}

\bibliography{custom}


\appendix
\input{sections/appendix}



\end{document}

%% file: sections/01-Introduction.tex
\section{Introduction}\label{sec:Intro}

\begin{figure}[t]
        \centering
	\includegraphics[width=1 \linewidth]{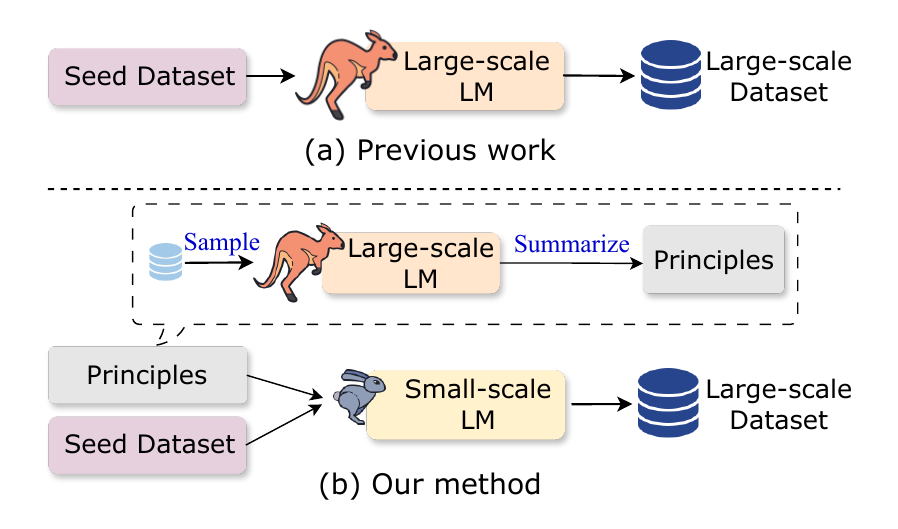}
        \caption{Comparison between existing self-instruct methods (a) and our proposed \model (b). Instead of using large-scale LM to generate an instruction-tuning dataset, \model employs small-scale LM to generate a high-quality dataset using the principles as guidance.}
 \label{fig:intro}
\end{figure}

Instruction tuning~\citep{humaninstruct,flant5} is a crucial step in training large language models (LLMs). Through supervised learning on a large dataset of instruction-response pairs, LLMs acquire the ability to follow instructions and utilize the knowledge accumulated during pre-training.
This capability allows LLMs to be applied to a wider range of tasks. 
Additionally, this training approach enables LLMs to be applied on more task-specific applications and more tasks in vertical domains.

Existing instruction tuning methods typically obtain large-scale datasets from two sources.
The first involves employing a large number of human annotators to annotate the training data \cite{lima,t0sanh}. 
While this method can produce high-quality datasets, it is expensive and time-consuming. 
And the second method uses a larger-scale LM to generate the instances, known as the self-instruct framework ~\citep{selfinstruct,alpaca}. 
This approach usually requires human annotation of a small seed dataset, after which a large-scale LM mimics the examples in the seed dataset and incorporates the domain knowledge to generate a larger dataset. 
This method has been widely applied to tasks such as question answering and reasoning in specific domains, \eg tool learning \cite{gao2024confucius}.

Since the self-instruct method requires a powerful large-scale LM to perform data annotation in order to obtain high-quality datasets, it involves providing these large-scale LMs with some task 
examples in the input and having them generate a large-scale dataset. 
Due to the large number of parameters in these LLMs, the instances generation process inevitably produces a significant amount of carbon emissions. 
However, using smaller LLMs to generate instances fails to meet the high-quality dataset requirements for instruction fine-tuning \cite{small}.

On the other hand, most self-instruct methods \cite{wizardlm2024xu,alpacagpt4,Coachlm} rely on closed-source LLMs, such as ChatGPT and Claude. 
During the instances generation process, we need to provide these closed-source models with example instances as well as additional knowledge required for instances generation \cite{li2024selfalignment}. 
For example, when annotating domain-specific tool-learning instances, we need to include internal tool API documentation in the prompts to generate data using these tools \cite{qin2024toolllm,shi2024chain}. 
However, this approach risks leaking internal tool documentation. 
In some applications, due to privacy and security concerns, we cannot have permission to transfer these tool documents to closed-source model service providers.

Therefore, in this paper, we propose the \fullmodel (\model) method. 
This approach involves having the large LM generate a few task-completion principles, which are then used by a smaller LM to generate a high-quality, large-scale dataset by following these principles. 
This method avoids having the large model directly generate the entire dataset and eliminates the need to transfer domain-specific knowledge to closed-source model service providers. 
First, we introduce the \first method, where the large-scale LM reflects and summarizes multi-level principles for completing tasks based on a seed dataset annotated by humans. 

The \first method starts by randomly sampling several subsets from the seed dataset. 
To generate more targeted task principles, we first let a smaller LM directly generate instances based on the seed dataset. 
Then, we use a large-scale LM to generate task principles that better describe the points where the smaller LM is prone to errors.
Subsequently, through further summarization and reflection, we obtain a refined, high-level task principle pool.
Next, we use the \second method to follow these task-completion principles and generate a large-scale instruction fine-tuning dataset by a smaller LM. 
This instances generation method also mitigates the high carbon emissions when directly using large-scale LMs to generate instances. 
Finally, this dataset is used for instruction fine-tuning training.

Experiments on several benchmark datasets reveal that the LLM trained on the dataset constructed by \model achieves comparable performance to those trained on datasets directly annotated by large-scale LMs. 
This phenomenon demonstrates the effectiveness of our principle-based instances generation method. 
Compared to directly using smaller-scale LMs to construct datasets, our principle-based method achieves consistently better performance. 
Additionally, we also explore the performance of variant types of datasets, the impact of different experiences on dataset quality, and a comparison of carbon emissions among various methods.

Our contributions are as follows:
\textit{(i)} We propose a principle-based method \model{} for constructing instruction tuning datasets, which reduces high carbon emissions and solves privacy leakage issues when using closed-source LLMs;
\textit{(ii)} We introduce the \first{} method to build a low-redundancy, high-quality pool of task principles;
\textit{(iii)} We propose the \second{} method to enable smaller-scale LLMs to follow task principles and generate high-quality instruction tuning datasets; and
\textit{(iv)} We conducted extensive experiments on several benchmark datasets, demonstrating that our \model achieves comparable performance to directly using large-scale LMs for annotation while significantly reducing carbon emissions.

%% file: sections/02-related-work.tex
\section{Related work}\label{sec:related-work}

Self-instruction~\cite{selfinstruct} has emerged as an effective method for utilizing LLMs to synthesize instruction fine-tune datasets. 
It starts from a set of manually written seed datasets and utilizes LLM to synthesize instances.
\texttt{Alpaca}~\cite{alpaca} propose to improve the performance of llama-7b through the distillation of a larger LLM, \eg \texttt{text-davinci-003}. 
Then, \texttt{Alpaca-GPT4}~\cite{alpacagpt4} achieve better performance by using the more powerful GPT-4 as the instance generation model.
Taking advantage of the rewriting capabilities of LLMs, WizardLM~\cite{wizardlm2024xu} iteratively employs ChatGPT to rewrite initial instructions into increasingly complex instructions.
For some specific domains such as math~\cite{wizardmath,commonmath_li2024}, tool-learning~\cite{qin2024toolllm,gao2024confucius} and dialogue~\cite{baize_chat_xu2023,enhancing_chat_ding2023}, generating synthetic dataset via self-instruct has also achieved significant success.

However, most of the self-instruct frameworks heavily rely on powerful closed-source LLMs, leading to environmental impact and privacy risks. 
Therefore, in this paper, we focus on enhancing the instances synthesis capabilities of open-source smaller LLMs.

%% file: sections/03-method.tex
\section{Task Definition}

The self-instruct approach~\cite{selfinstruct} typically involves the following steps: 
First, we employ human annotators to construct a seed dataset $\mathcal{D}_{seed}$.
Then $\mathcal{D}_{seed}$ is used as in-context examples for a large LM $\mathcal{M}_{L}$, with prompts designed to guide $\mathcal{M}_{L}$ to generate a larger dataset $\mathcal{D}_{t}$ consistent with $\mathcal{D}_{seed}$.
Finally, the large dataset $\mathcal{D}_{t}$ is used to fine-tune a model $\mathcal{M}_{t}$.

In \model, we also use a small seed dataset $\mathcal{D}_{seed}$ as in-context examples for a large language model $\mathcal{M}_{L}$. 
However, instead of directly generating a large-scale dataset by prompting the large LM $\mathcal{M}_{L}$, we design prompts to guide $\mathcal{M}_{L}$ to generate a set of task-completion principles $\mathcal{P}$. 
Then, we use the seed dataset $\mathcal{D}_{seed}$ and the task-related principles $\mathcal{P}$ as prompts to a smaller language model $\mathcal{M}_{g}$, which generates a dataset $\mathcal{D}_{t}$ similar to $\mathcal{D}_{seed}$ under the guidance of $\mathcal{P}$. 
Finally, the dataset $\mathcal{D}_{t}$ is used to fine-tune a model $\mathcal{M}_{t}$.

\section{\model Method}\label{sec:Method}

\begin{figure*}[htbp]
        \centering
	\includegraphics[width=1\linewidth]{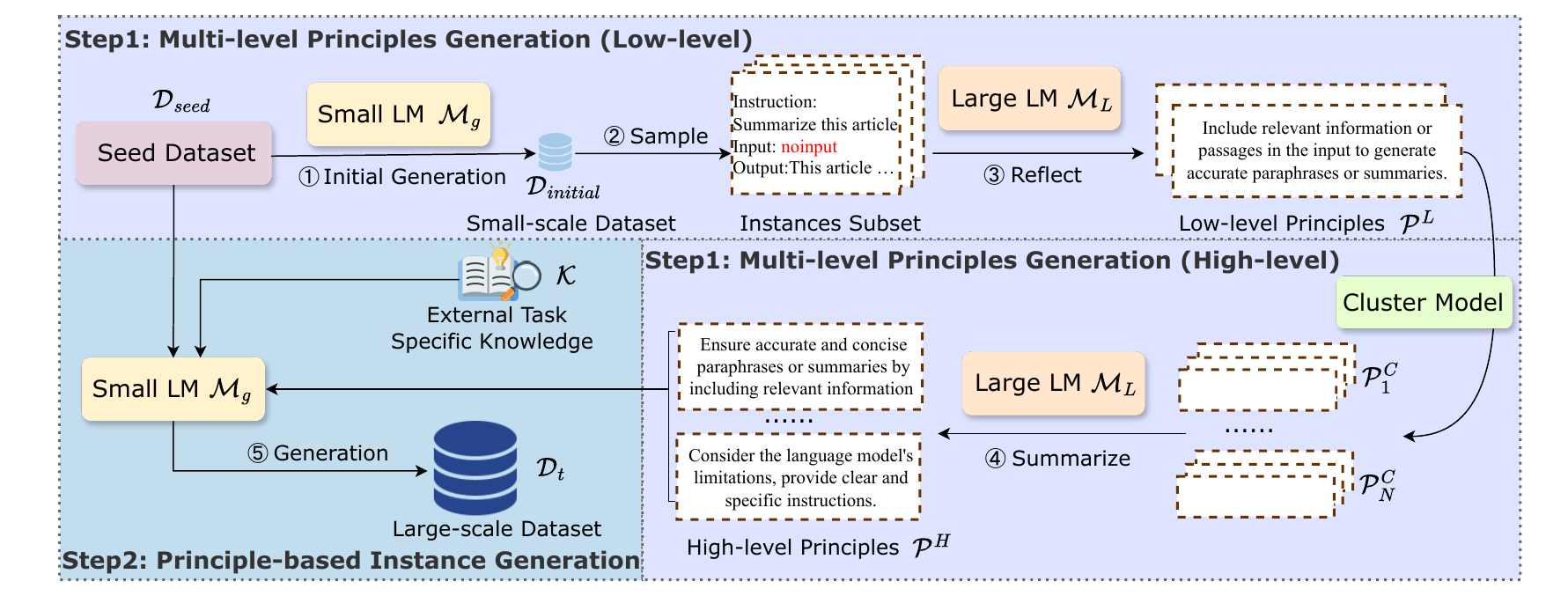}
        \caption{
        The architecture of our proposed \model. The \first consists of two parts: First, we employ large LM $\mathcal{M}_L$ to generate low-level principles. Second, these low-level principles will be clustered and summarized by $\mathcal{M}_L$ into high-level principles $\mathcal{P}^H$.
        The \second uses a smaller LM $\mathcal{M}_g$ to generate instances guided by high-level principles $\mathcal{P}^H$ and external knowledge $\mathcal{K}$.}
       
        \label{fig:method}
\end{figure*}

In our \model method, there are two main steps. 
First, we use the \first method, where the large-scale language model $\mathcal{M}_{L}$ reflects and summarizes a set of task-related principles $\mathcal{P}$ based on the given seed dataset $\mathcal{D}_{seed}$. 
Then, under the guidance of $\mathcal{P}$, we employ the \second method to use a smaller language model $\mathcal{M}_{g}$ to generate a large-scale dataset $\mathcal{D}_{t}$ similar to $\mathcal{D}_{seed}$. 
The parameters of the language model $\mathcal{M}_{g}$ is significantly smaller than that of the large-scale language model $\mathcal{M}_{L}$ used for generating the principles $\mathcal{P}$. 
Thus, our method can significantly reduce carbon emissions.

\subsection{\first}

To further reduce reliance on manual data annotation, we first use the language model $\mathcal{M}_{g}$ to perform a simple expansion of the manually annotated seed dataset $\mathcal{D}_{seed}$, resulting in an expanded dataset $ \mathcal{D}_{initial} $:
\begin{equation}
    \mathcal{D}_{initial} = \mathcal{M}_{g} (\mathcal{D}_{seed}, \mathcal{K}, \mathcal{I}_s),
\end{equation}
where $ \mathcal{K} $ represents the external knowledge required for instances generation, and $ \mathcal{I}_s $ is the prompt used to guide the language model $\mathcal{M}_{g}$ in instances augmentation. 
It is important to indicate that although $ \mathcal{D}_{initial} $ includes some external knowledge, the dataset remains relatively small ($|\mathcal{D}_{seed}| < |\mathcal{D}_{initial}| \ll |\mathcal{D}_{t}|$). 
This allows us to filter instances or anonymize the knowledge within $ \mathcal{D}_{initial} $, thus mitigating the risk of privacy leakage.

Subsequently, to obtain more diverse task-completion principles, we perform multiple random samplings on the dataset $ \mathcal{D}_{initial} $.
Then, we obtain several subsets $\{d_1, d_2, \dots, d_T\}$, where each subset $ d_i $ contains multiple task instances, and $T$ is the number of subsets. 
We then use the large-scale language model $\mathcal{M}_{L}$ to reflect on each subset $ d_i $ and summarize \textbf{low-level principle} $ \mathcal{P}^L_i $:
\begin{equation}
\mathcal{P}^L_i = \mathcal{M}_{L}(d_i, \mathcal{I}_p), \label{equ:low-principle}
\end{equation}
where $ \mathcal{P}^L_i $ includes several natural language descriptions of task principles, and $ \mathcal{I}_p $ is the prompt guiding the language model $\mathcal{M}_{L}$ to generate the low-level task principles:
\begin{tcolorbox}[colback=black!1!white,colframe=black!57!white,boxsep=1pt,left=1pt,right=1pt,top=1pt,bottom=1pt]
\instrctionsize
<INSTRUCTIONS DATA>\\
Conduct a thorough analysis of the given instructions output pairs. Provide clear principles that can be derived from this analysis to improve future and outputs. We are not focused on this one data point, but rather on the general principle.
\end{tcolorbox}
Thus, we obtain the set of low-level task principles for each subset, $ \mathcal{P}^L = \{\mathcal{P}^L_1, \mathcal{P}^L_2, \dots, \mathcal{P}^L_T\} $.

Intuitively, some low-level task principles may be redundant or overly specific.
To address this, we first vectorize all low-level task principles $ \mathcal{P}^L $ using a semantic embedding model.
Then, we apply a clustering algorithm to these semantic representations, resulting in $ N $ clusters of task principles, $ \mathcal{P}^C = \{\mathcal{P}^C_1, \mathcal{P}^C_2, \dots, \mathcal{P}^C_N\} $. 
Specifically, we use a soft clustering algorithm that doesn't require a pre-defined number of clusters.

To reduce computational load and further reflect on the task principles, we use the language model $\mathcal{M}_{L}$ to summarize each task principle cluster to construct the \textbf{high-level principle}:
\begin{equation}
\mathcal{P}^H_i = \mathcal{M}_{L}(\mathcal{P}^C_i, \mathcal{I}_h),
\end{equation}
where $ \mathcal{P}^H_i $ is a high-level task principle, and $ \mathcal{I}_h $ is the prompt guiding the language model $\mathcal{M}_{L}$ to generate high-level task principles: 
\begin{tcolorbox}[colback=black!1!white,colframe=black!57!white,boxsep=1pt,left=1pt,right=1pt,top=1pt,bottom=1pt]
\instrctionsize
<LOW LEVEL PRINCIPLES>\\
Create a high level and insightful principle to improve future responses.  Focus on capturing the essence of the feedback while eliminating redundancies.
\end{tcolorbox}
Then, we obtain the high-level task principle set $ \mathcal{P}^H = \{\mathcal{P}^H_1, \mathcal{P}^H_2, \dots, \mathcal{P}^H_N\} $, containing $ N $ high-level principles necessary for completing the tasks.

\subsection{\second}

Generating a large-scale instruction fine-tuning dataset is the most resource-intensive part of the self-instruction process. 
Therefore, after obtaining these task principles, to reduce reliance on the large-scale model $\mathcal{M}_{L}$, we use a smaller parameter model $\mathcal{M}_{g}$ to generate a large amount of instruction fine-tuning data:
\begin{equation}
d = \mathcal{M}_{g}(\mathcal{P}^H, \mathcal{K}, \mathcal{I}_g),\label{eq:gen_large_data}
\end{equation}
where $ d $ is a generated instruction tuning instance, and $\mathcal{I}_g$ is the prompt guiding model $\mathcal{M}_{g}$ to generate instances based on the task principles $ \mathcal{P}^H $:
\begin{tcolorbox}[colback=black!1!white,colframe=black!57!white,boxsep=1pt,left=1pt,right=1pt,top=1pt,bottom=1pt]
\instrctionsize
You are asked to come up with a set of 20 diverse task instructions.\\
The following insights and guidelines may improve responses:\\
<PRINCIPLES>

\end{tcolorbox}

By repeatedly executing Equation~\ref{eq:gen_large_data}, we can obtain a large-scale instruction fine-tuning dataset $\mathcal{D}_{t}$, where $|\mathcal{D}_{t}| \gg |\mathcal{D}_{seed}|$.
Since $\mathcal{M}_{g}$ has significantly fewer parameters than model $\mathcal{M}_{L}$, the carbon emissions during the generation of a large dataset are greatly reduced. 
Additionally, model $\mathcal{M}_{L}$ only receives the small-scale dataset $ \mathcal{D}_{initial} $ as in-context learning examples throughout the process and does not use our external knowledge $ \mathcal{K} $. 
This also prevents the leakage of the private knowledge base $ \mathcal{K} $ to external model service providers.
Finally, we use the dataset $\mathcal{D}_{t}$ to fine-tune a language model $\mathcal{M}_{t}$.

%% file: sections/04-experiment.tex
\section{Experimental Setup}\label{sec:Experiment}

\subsection{Datasets}

To verify the effectiveness of our proposed \model, we use the generated instruction tuning dataset to fine-tune a model $\mathcal{M}_{t}$, and then evaluate the model performance of $\mathcal{M}_{t}$ using the following benchmark datasets. 
We classify the datasets into four categories according to their task definition:

\header{Truthfulness and Knowledge Task}
\textit{TruthfulQA} is a popular benchmark used to test the model's hallucination~\cite{truthfulqa}. 
We utilized it to gauge the truthfulness of models, reporting accuracy under a 0-shot setting.
And we use \textbf{MMLU}~\citep{MMLU} to assess the factual knowledge of models and report the mean accuracy under a 5-shot setting.

\header{Commonsense Reasoning Task} 
\textit{WinoGrande} is to choose the right option for a given sentence formulated as a fill-in-a-blank~\cite{winogrande2021}.
\textit{GPQA}~\cite{gpqarein} is a multiple-choice written by domain experts. Both of them require the model to have commonsense reasoning abilities. We report the accuracy of these two datasets under 5-shot and 3-shot settings. 

\header{Code Generation Task} 
We test the programming capabilities using \textit{HumanEval}~\cite{humanevalchen} and perform the decoding using two different temperatures: 0 and 0.8. 
We report the better Pass@10 from these two decoding results.

\header{Math Reasoning Task} 
We use the \textit{GSM8K} dataset~\cite{gsm8kcobbe} to evaluate the ability of solving multi-step mathematical reasoning problems and report the exact match under the 5-shot setting.

\subsection{Baselines}

We compare our method with the following widely-used self-instruction methods:
\textit{(i)} \textbf{Alpaca} \citep{alpaca} is 52K instruction-tuning dataset generated by text-davinci-003 using the self-Instruct technique; 
\textit{(ii)} \textbf{Alpaca-GPT4} \citep{alpacagpt4} follows the same methodology as Alpaca, but incorporates GPT-4 as its teacher model.
\textit{(iii)} \noindent\textbf{WizardLM} \citep{wizardlm2024xu} uses Evol-Instruct to rewrite Alpaca dataset step by step into more complex instructions; and 
\textit{(iv)} \textbf{Alpagasus} \citep{chen2024alpagasus} uses the powerful GPT-3.5 (\textit{as known as} ChatGPT) to score and select 9K instances from the original Alpaca dataset.

\subsection{Implementation Details} \label{sec:implementation}

We employ two LLMs as the instances generation model $\mathcal{M}_{g}$ to verify the generalization of our proposed method: Zephyr-7B-Beta~\cite{zephyr} and Llama-3-8B-Instruct~\cite{llama3modelcard}.
And we also use the generated dataset to fine-tune two language models $\mathcal{M}_{t}$, i.e., Qwen-1.8B~\cite{qwen2} and Gemma-2B~\cite{gemma_2024}.

We construct several instruction tuning datasets, each containing 20K examples, using our proposed \model method and several other baseline methods (\eg Alpaca). We set the number of subsets $T$ to 10, and the size of each subset, $|d_i|$, to 10.

All our experiments are conducted on 4 $\times$ NVIDIA A800 (80GB) GPUs, with the same setting and hyperparameters of Alpaca.
We employ gpt-3.5-turbo from OpenAI as the $\mathcal{M}_L$ to reflect and summarize. The generated principles are presented in Appendix ~\ref{sec:principles}. And we also present a comparison between these principles and those written by humans in Appendix ~\ref{app:human_written}.

%% file: sections/05-results.tex
\section{Experimental Results}\label{sec:result}
\input{table/main}
\subsection{Overall Performance}
Table~\ref{tab:main} shows the evaluation results of our proposed \model and other self-instruct baselines.
Compared to directly using large LMs for instances generation (\eg \texttt{Alpaca w/ GPT4}), our \model w/ Zephyr-7B achieves comparable performance on two backbone models (including Gemma-2B and Qwen-1.8B). 
On the GPQA and Winogrande datasets, it even outperformed models that used GPT-4 for instance generation.

From Table~\ref{tab:main}, it can be seen that although the performance of \model is slightly lower than that of models trained with the dataset generated by GPT-4, it outperforms models that directly use smaller LMs to generate instances (\eg Zephyr) on most of the datasets. 
We can find that compared with the \texttt{Alpaca w/ Zephyr} which directly employs the Zephyr as instances generation model $\mathcal{M}_{g}$, our proposed \model achieves consistent improvement when using different backbone LLMs (\eg Gemma-2b and Qwen-1.8B).
This indicates that the task principles effectively help these smaller LMs generate a higher-quality instruction-tuning dataset.
To demonstrate the generalization ability, we also conduct experiments using Llama-3 as the instance generation LLM.
The \model w/ Llama3 outperforms the \texttt{Alpaca w/ Llama3} in the majority of datasets.

We also find that the improvements of \model are not consistent across different types of datasets. 
In most knowledge-centric datasets (\eg MMLU), the performance improvement of our model compared with directly using smaller models for instance generation is not as high as it is for other types of datasets. 
It is intuitive that task principles usually only provide higher-level guidance on methodology or instance formats for completing tasks, whereas knowledge-based datasets require extensive knowledge content as support, which is not included in the task principles. 
For logical reasoning tasks (\eg GPQA), the problem-solving experience included in the task principles can effectively help the model generate high-quality data. 
Therefore, our \model achieves better improvements than the baselines without principles on logical reasoning tasks.

\subsection{Analysis of Carbon Emission}

\input{table/carbon}

Since one of the motivations of our proposed \model is to reduce carbon emission when generating large-scale instruction tuning dataset, we compare the carbon emission between \model w/ Zephyr-7B and \texttt{Alpaca w/ GPT4} as instances generation LLM.
The detailed calculation method of carbon emission is illustrated in Appendix~\ref{app:carbon}.
Additionally, we directly report the number of LLM tokens used by different instances generation LLMs. 
Table~\ref{tab:open} shows a comparison between these two methods. 
From Table~\ref{tab:open}, it is evident that our \model method significantly reduces carbon emissions when generating large-scale instances (with p-value < 0.05), which demonstrates that \model is greener than directly using large LM to generate instances.

\input{table/ablation}
\subsection{Ablation Study}

To verify the effectiveness of each module in our \model, we employ three variant models that remove each module and conduct experiments on each type of dataset.
\textit{w/o initial} indicates no initial generation, using $\mathcal{D}_{seed}$ to replace with $\mathcal{D}_{initial}$. \textit{w/o sample} means there are no samples, and $\mathcal{M}_L$ evaluates as much instances as possible within its context length. \textit{w/o cluster} signifies no clustering, utilizing a set of low-level principles to guide generation.

Table~\ref{tab:ablation1} shows the performance of each ablation model, and we can find that removing any module results in a certain degree of performance decline, which validates the necessity of each module.

\subsection{Analysis of External Knowledge}

Our method allows for the integration of external knowledge bases, which can further enhance performance in scenarios that require designing for user privacy, such as medical and tool learning. In this section, we investigate whether incorporating external knowledge can further improve model performance in medical settings \cite{li2023chatdoctor}. 
Specifically, we utilize medical records from HealthCareMagic as the external knowledge source to improve the capability of Zephyr-7B in generating doctor-patient interaction instances. Then, we use these instances to train a medical chatbot based on Gemma-2B and report its performance in Table~\ref{tab:knowledge}.
The results demonstrate that by integrating external knowledge, small-scale LMs can further improve its ability.

\input{table/knowledge}

\subsection{Effect of Different Principles}\label{sec:exp:radar}

\begin{figure}[t]
        \centering
	\includegraphics[width=1\linewidth]{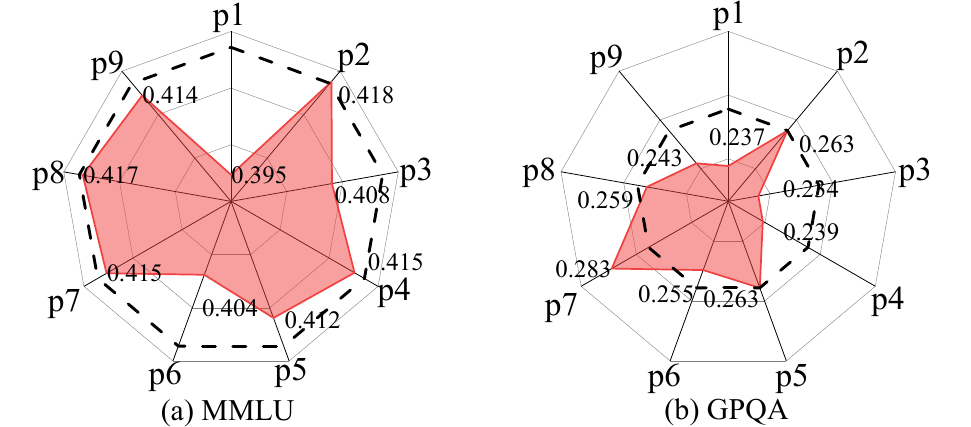}

        \caption{Effectiveness evaluation of each principle. Each vertex of the radar chart represents the performance of the model trained after removing the principle from the set. The dashed circle indicates the performance of the model using the whole set. }
 \label{fig:effect-principle}
\end{figure}

In the \model, we employ the $\mathcal{M}_{L}$ to generate several task principles to guide the $\mathcal{M}_{g}$ to generate a large-scale dataset.
Therefore, an intuitive question arises: \textit{Is each generated principle useful for improving the quality of the dataset? }
In this section, we individually remove each principle, then use the remaining set of principles to generate data, and finally fine-tune the model.
We use Zephyr-7B as $\mathcal{M}_g$ and Gemma-2B as $\mathcal{M}_t$.
Figure~\ref{fig:effect-principle} shows the performance changes of the model on two datasets after removing each principle. 
From Figure~\ref{fig:effect-principle}, we can find that the performance declines compared to using the full set of principles after removing any principle, which indicates the necessity of these principles. 
Additionally, we also found that some principles are crucial for instances quality. 
For example, we show the principle $ \mathcal{P}^H_1 $ from the MMLU and GPQA dataset:
\begin{tcolorbox}[colback=black!1!white,colframe=black!57!white,boxsep=1pt,left=1pt,right=1pt,top=1pt,bottom=1pt]
\instrctionsize
Ensure clear and concise communication by aligning the output with the prompt, maintaining proper formatting and grammar, addressing the task accurately and comprehensively, and organizing the output logically for better readability.
\end{tcolorbox}
This principle describes a critical method of generating data instances, which enables the model to generate comprehensive and detailed outputs aligned with the instruction.
Thus, this experiment demonstrates that our proposed \model can generate valuable task principles, thereby helping the model generate high-quality datasets.
We show more results and discussions on principle selection and data representation in Appendix~\ref{app:additional_exp}.

\subsection{Analysis of Dataset Size}

\begin{figure}[t]
        \centering
	\includegraphics[width=1 \linewidth]{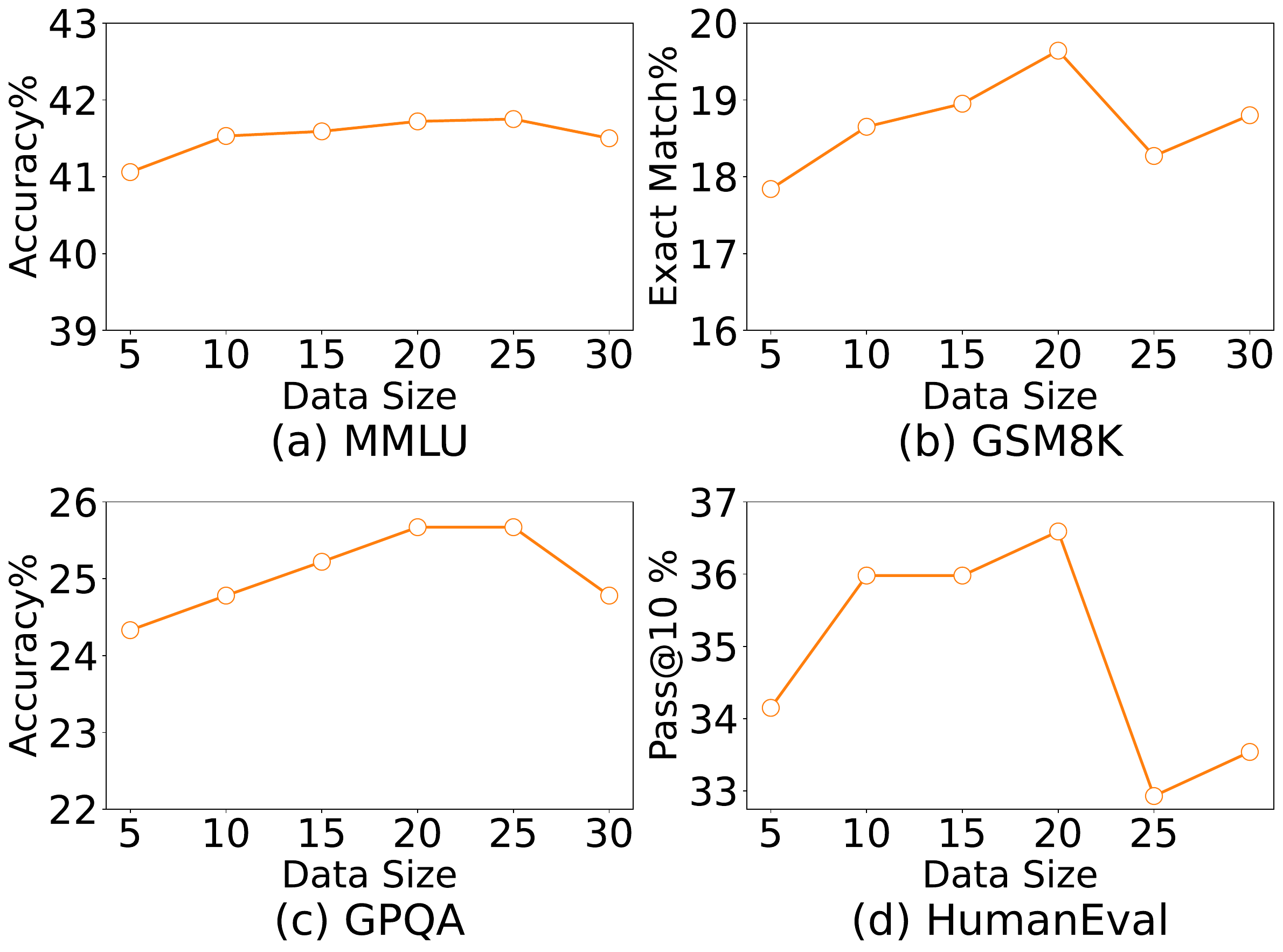}
        \caption{Model performance of using different scales.}
 \label{fig:size}
\end{figure}

In the experiments shown in Table~\ref{tab:main}, all methods generate 20K samples for model training. 
In this section, we explore the impact of generating datasets of different sizes. 
Figure~\ref{fig:size} shows how model performance changes when using datasets of different sizes generated by Zephyr-7B, while keeping the number of task principles constant.
From Figure~\ref{fig:size}, we can find that as the dataset size increases from 5K to 20K, the performance of our model gradually improves. 
However, when the dataset size exceeds 20K, the performance no longer increases. 
The reason for this phenomenon is evident that with a fixed number of task principles, a dataset with enough samples is sufficient for the model to fully grasp these principles, while an excessively large dataset leads to the model overfitting these principles.

\subsection{Analysis of Different Backbone LLM}

\begin{figure}[t]
        \centering
	\includegraphics[width=1 \linewidth]{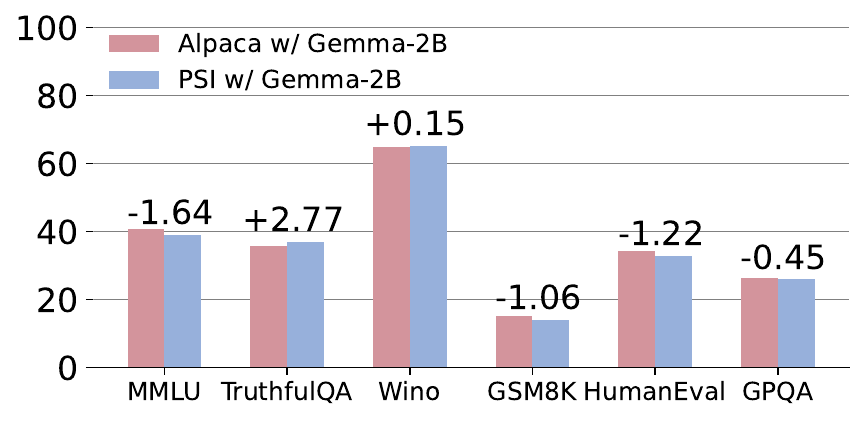  }
        \caption{Comparison between \model and Alpaca which both use a small-scale LM as the instance generation model.}
 \label{fig:gemma_gemma}
\end{figure}

In our experiments, we utilized the 7B model for generating instances as the $\mathcal{M}_{g}$, while a smaller 2B model was employed for fine-tuning. 
An intuitive experiment is to see whether a smaller model with 2B parameters could effectively generate instances. 
In this section, we replace the $\mathcal{M}_{g}$ in \model 

with  Gemma-2B and use the generated dataset to finetune another Gemma-2B model. 

Figure~\ref{fig:gemma_gemma} illustrates the comparison results. 
It shows that the performance of the model trained with instances generated by the smaller Gemma-2B is comparable with the model trained with \texttt{Alpaca}, which does not use any principles. 
This outcome indicates that the task principles did not help the model generate higher-quality data.
An intuitive reason is that the smaller model cannot understand the complex instructions (\eg following task principles), leading to lower-quality datasets.
Conversely, the results in Table~\ref{tab:main} demonstrate that the instances generation LLM with 7B parameters can understand complex instructions and achieves better performance than \texttt{Alpaca} without using principles.

\subsection{Analysis of Principle Number}

\begin{figure}[t]
        \centering
	\includegraphics[width=1 \linewidth]{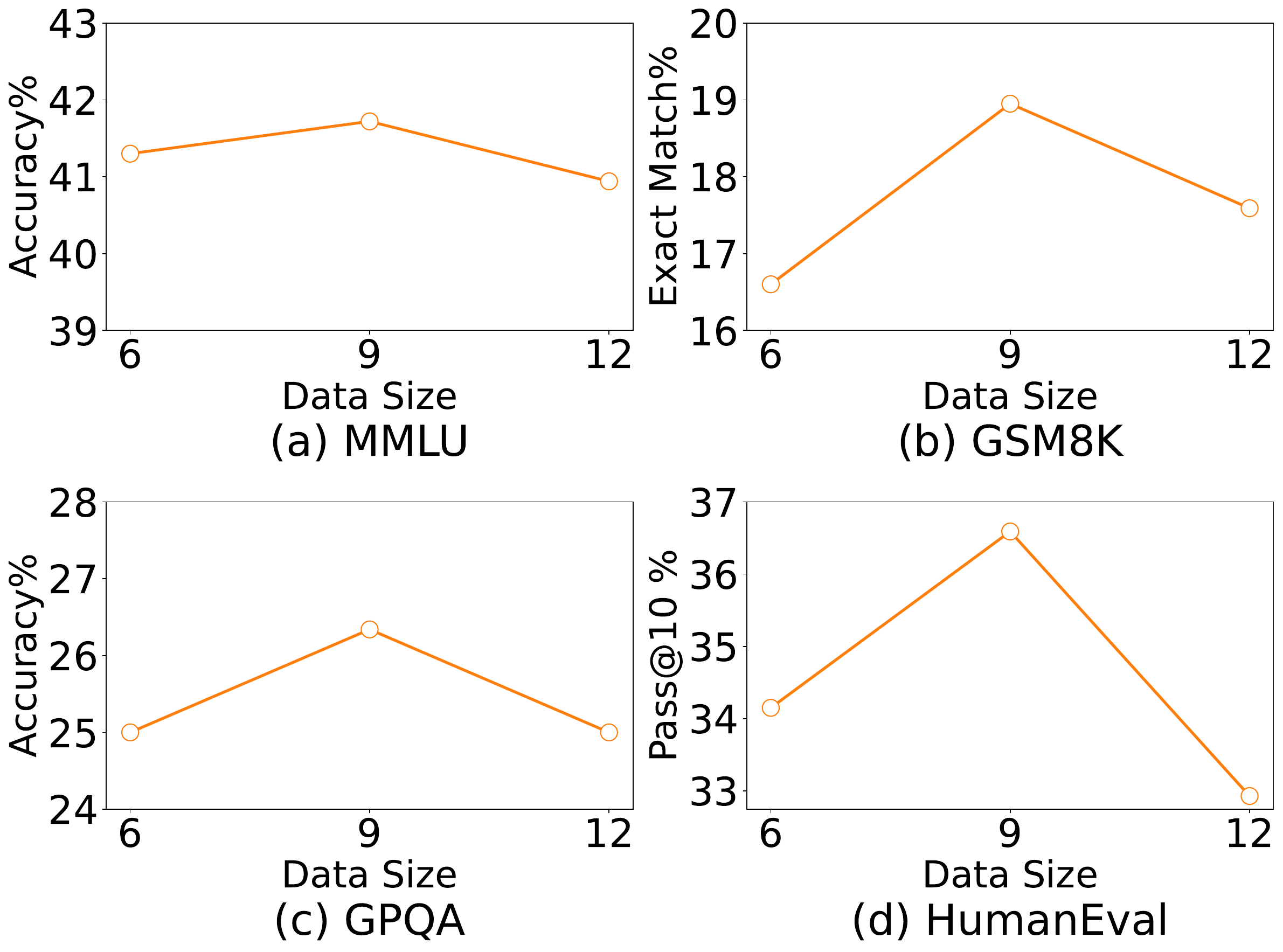}
        \caption{The model performance of using the different number of principles.}
        \label{fig:principle_num}
\end{figure}

In Equation~\ref{equ:low-principle}, we randomly sample a subset $d_i$ from the expanded dataset $\mathcal{D}_{initial}$
and then use the LLM $\mathcal{M}_L$ to summarize the low-level task principle $\mathcal{P}^L_i$.
Multiple subsets $d_i$ are sampled to construct the principles. 
In this section, we analyze the effect of the number of principles.
As shown in Figure~\ref{fig:principle_num}, we can find that the model achieves the best performance when we sample 10 times.
Additionally, neither increasing nor decreasing the number of samples can improve performance. 
The reason for this phenomenon is that too few times of samples may not encompass the experience needed to complete the task, while too many samples can result in redundant principles confusing the $\mathcal{M}_g$.

\subsection{Human Evaluation}

Since the instruction following task aims at generating an open-ended response as the answer, we conduct a human evaluation to qualitatively evaluate the performance of our \model.
We randomly sample 40 instructions from each dataset: Dolly~\cite{dolly2023}, Koala~\cite{koala_blogpost_2023}, and Vicuna~\cite{vicuna2023}, and use the fine-tuned model to generate the response for each instruction.
In this experiment, we employ the Zephyr-7B as instance generation model $\mathcal{M}_{g}$, and use Gemma-2B as the fine-tuned model $\mathcal{M}_{t}$ for both \texttt{Alpaca} and \model.
We employ 3 highly educated human annotators to evaluate the generated responses. The detailed evaluation criteria is provided in Appendix~\ref{sec:app:human}.

\begin{figure}[t]
    \centering
    \includegraphics[width=1 \linewidth]{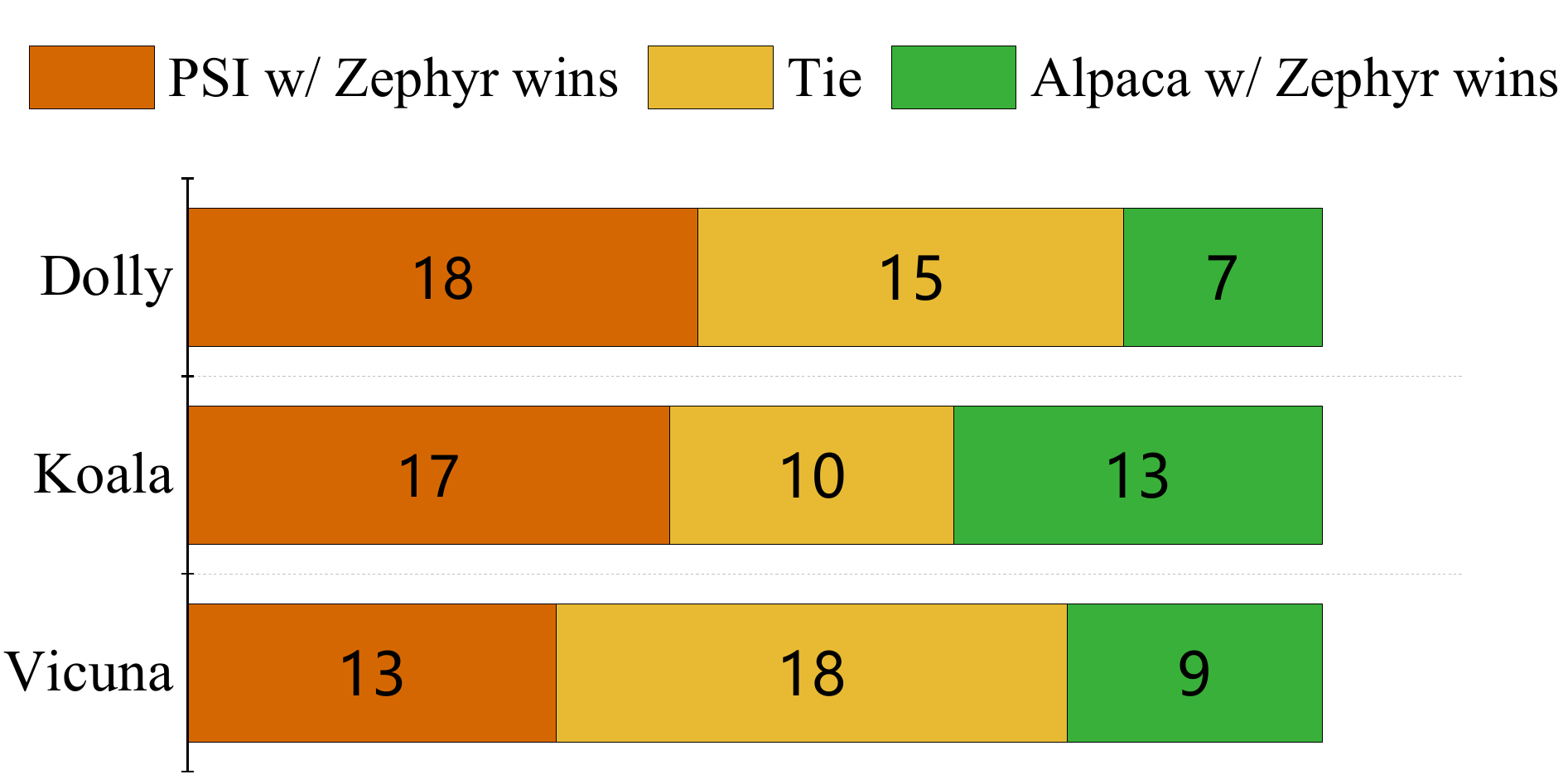}
    \caption{Human evaluation of the model trained by different self-instruct methods.}
    \label{fig:human_eval}
\end{figure}

The Cohen's kappa for the human annotators is 0.56, 
indicating moderate inter-annotator agreement.
As shown in Figure~\ref{fig:human_eval}, on three datasets, the models trained on datasets generated by \model achieved better performance.

%% file: table/main.tex
\begin{table*}[ht]
\centering
\resizebox{1.0\textwidth}{!}{
\begin{tabular}{l cc ccccc  cc  }
\toprule

\multirow{2}{*}{\bf Method} &\bf Instance Gen LLM&  \multirow{2}{*}{\textbf{MMLU}}&   \multirow{2}{*}{\textbf{TruthfulQA}}& \multirow{2}{*}{\textbf{GPQA}} & \multirow{2}{*}{\textbf{Winogrande}}& \multirow{2}{*}{\textbf{GSM8K}} &  \multirow{2}{*}{\textbf{HumanEval}}& \multirow{2}{*}{\textbf{Overall}}   \\

    & $\mathcal{M}_g$ &    &  &  &   &  &    \\
\midrule
\rowcolor{gray!40}
\multicolumn{1}{l}{ \it Base Model ($\mathcal{M}_t$): Gemma-2B}  & - & 40.7 & 33.2 & 23.4 & 65.1   & 18.5   & 34.2     & 35.9   \\

~~~ Alpaca & text-davinci-003 & 38.7 &   38.2  & 24.6 & 66.9   & 13.3   & 35.4    &36.2 \\

~~~ AlpaGasus & gpt-3.5-turbo & 41.6 & 36.3  &27.0 & 66.5   &  15.5   & 37.8   & 37.5 \\

~~~ WizardLM  & gpt-3.5-turbo   & 41.0 & 43.2  & \textbf{27.5} & 64.9  & 19.4  & \textbf{44.5}  & \textbf{40.1}    \\

~~~ Alpaca & GPT-4   & \textbf{42.3} & \textbf{43.6} & 25.7 & 65.8  & 20.2  & 37.2    & 39.1  \\ 

~~~ Alpaca & Zephyr-7B     &  40.4 & 42.0 & 25.0 & 65.8 & 17.3   & 31.7    &37.0  \\

~~~ \model &  Zephyr-7B & 41.7$_{\uparrow 1.3}$ &  43.5$_{\uparrow 1.5}$ & 26.3$_{\uparrow 1.3}$ &  66.1$_{\uparrow 0.3}$     & 19.0$_{\uparrow 1.7}$ & 36.6$_{\uparrow 4.9}$    & 38.9 $_{\uparrow 1.9}$  
\\


~~~ Alpaca & Llama3-8B   & 40.9 &   42.9 & 25.7 & 65.7  & 19.4   & 35.4    &38.3 \\
~~~ \model &  Llama3-8B    & 39.9$_{\downarrow 1.0} $  & 41.8$_{\downarrow 1.1}$ & 26.6$_{\uparrow 0.9}$ & \textbf{66.9} $_{\uparrow 1.2}$   & \textbf{22.3}$_{\uparrow 2.9}$ & 37.8$_{\uparrow 2.4}$    & 39.2$_{\uparrow 0.9}$ 
\\

\midrule
\rowcolor{gray!40}
\multicolumn{1}{l}{ \it Base Model ($\mathcal{M}_t$): Qwen-1.8B} &  -  & 45.3 & 39.4& 27.2 & 61.6   &34.5    &32.9  & 40.2   \\ 

~~~ Alpaca & text-davinci-003 & 45.3  &   37.5 & 28.4 & 61.3   & 17.7   & 34.8     & 37.5 \\

~~~ AlpaGasus & gpt-3.5-turbo       & 46.5 & 38.5 & 27.5 & 61.6   &  26.9   & \textbf{36.0}     & 39.5   \\

~~~ WizardLM & gpt-3.5-turbo    & 46.3 & 39.9 & \textbf{29.2} & 62.5  & \textbf{36.6}  & 25.0    & 39.9    \\

~~~ Alpaca & GPT-4        & 46.2 & \textbf{42.3} & 28.4 & 60.9  & 34.4  & 34.8     & \textbf{41.2} \\

~~~ Alpaca & Zephyr-7B      & 45.6 & 41.6 & 25.5 & 62.2 & 29.4  & 27.4    & 38.6 \\

~~~ \model & Zephyr-7B        & \textbf{46.9}$_{\uparrow 1.3}$ & 41.3$_{\downarrow 0.3}$ & 28.8$_{\uparrow 3.3}$& \textbf{63.3}$_{\uparrow 1.1}$   & 31.0$_{\uparrow 1.6}$ & 29.3$_{\uparrow 1.9}$    & 40.1$_{\uparrow 1.5}$ \\


~~~ Alpaca & Llama3-8B    & 45.7 & 36.1  & 28.1& 61.1 & 31.9   & 26.8    &38.3 \\
~~~ \model & Llama3-8B       & 45.8$_{\uparrow 0.1}$ & 40.4$_{\uparrow 4.3}$ & 29.0$_{\uparrow 0.9}$ & 62.1$_{\uparrow 1.0}$    & 33.1$_{\uparrow 1.2}$ & 35.4$_{\uparrow 8.6}$    &41.0 $_{\uparrow 2.7}$
\\

\bottomrule
\end{tabular}
}
\caption{The performance of two base models: Gemma-2B and Qwen-1.8B, which are trained on the instruction tuning datasets generated by our proposed \model and baselines respectively.}
\label{tab:main}
\end{table*}

%% file: table/carbon.tex
\begin{table}[t]
  \centering
\resizebox{1.0\columnwidth}{!}{
    \begin{tabular}{lccc}
    \toprule
    \multirow{2}[0]{*}{\textbf{Method}} & \multicolumn{2}{c}{\textbf{Token Consume}} & \textbf{Carbon Emission} \\
    \cmidrule(lr){2-3}
    
    & \multicolumn{1}{c}{$\mathcal{M}_{L}$ (\textbf{GPT})} & \multicolumn{1}{c}{$\mathcal{M}_{g}$ (\textbf{Zephyr})}  & \textbf{(kgCO$_{2e}$)} \\
    \midrule
    \texttt{AlpaGasus} & 5,345,600  &  -   &   40.72  \\
    \texttt{WizardLM} & 7,834,077  &  -   & 34.74    \\
    \cbkgrnd \texttt{Alpaca w/ GPT4} &    \cbkgrnd 3,033,669   & \cbkgrnd -    &   \cbkgrnd 1.74        \\
    \model & 18,264  &  3,934,321   & 0.49\dubbelneer    \\

    \bottomrule
    \end{tabular}%
    }
  \caption{Comparison between \model and baseline methods in terms of token consumption and carbon emission.}
  \label{tab:open}
\end{table}%

%% file: table/ablation.tex
\begin{table}[!t]
\centering
\small
    \setlength\tabcolsep{3pt}

\begin{tabular}{@{}l cc  cc  @{}}

\toprule

\textbf{Method}

& MMLU & GSM8K & HumanEval & GPQA
\\
\midrule
\model
& 41.7 & 19.0 & 36.6 & 26.3
\\

~- \textit{w/o initial} 
& 41.6 &  17.7 & 34.8 & 26.1
\\

~- \textit{w/o sample}
& 41.5 &  18.6 & 30.5 & 24.8

\\
~- \textit{w/o cluster}
& 40.7 & 18.7 & 31.1 & 25.0

\\
\bottomrule
\end{tabular}
\caption{Ablation study on four types of datasets.}
\label{tab:ablation1}
\end{table}

%% file: table/knowledge.tex
\begin{table}[!t]
\centering
\small
    \setlength\tabcolsep{3pt}

\begin{tabular}{@{}l cc  cc  @{}}

\toprule

\textbf{Method}

& BLEU & ROUGE-L & BERTScore 
\\
\midrule



~ \texttt{Alpaca}
& 1.22 &  14.43 & 7.02
\\
~ \model  
& 1.28 &  14.71 & 7.23 
\\

~ \model w/ Knowledge
& \bf{1.31} &  \bf{14.85} & \bf{7.71}




\\
\bottomrule
\end{tabular}
\caption{The model performance using the external knowledge base in medical settings.}
\label{tab:knowledge}
\end{table}

%% file: sections/07-conclusion.tex
\section{Conclusion}

In this paper, we propose \fullmodel (\model) which is an environmentally friendly self-instruction framework that also avoids inputting task-related knowledge into proprietary LLMs, thereby protecting data privacy.
Specifically, we first proposed the \first method, which uses a large-scale LM to generate the task principles based on a given seed dataset. 
Then, by leveraging these task principles, a smaller-scale LM can be used to generate high-quality instruction-tuning datasets. 
We conducted extensive experiments on four types of instruction-following datasets, using various combinations of backbone LLMs. 
These experiments consistently demonstrated the effectiveness of our proposed \model. 
Additionally, compared to directly using large-scale LMs (\eg GPT-4) for instances generation, our method significantly reduces carbon emissions while maintaining comparable performance.

%% file: sections/limitations.tex
\section*{Limitations}~\label{limitations}
In this paper, we have conducted extensive experiments on four types of datasets, including knowledge, reasoning, coding, and math.
Due to the limited space, some task-specific datasets (\eg tool learning and protein structure analysis) are not included in our experiments.
However, these datasets have some task-specific features that our general framework does not include. 
We will expand our \model to many other fields in our future work.

\section*{Ethical Considerations}~\label{ethical}
In this paper, we propose the \model for instruction-tuning data generation based on LLMs. 
Although the LLMs have finished the alignment training, this method cannot entirely prevent generating unsafe data. 
However, since the \model generates data based on a small set of high-level task principles, we can constrain these task principles to minimize the generation of unsafe data. 
Additionally, to ensure complete safety, a manual review of the data is still required. 
Nevertheless, compared to manually constructing instruction-tuning datasets, only reviewing data significantly reduces the workload of annotators.

%% file: sections/appendix.tex
\section{Appendix}~\label{appendix}

\subsection{Prompts}\label{sec:prompt:system-prompt}

We provide our prompts used to generate principles. The prompt to generate low-level principles is shown in Table \ref{tab:prompt-low}. The prompt to generate high-level principles is shown in Table \ref{tab:prompt-high}. The prompt guiding $\mathcal{M}_g$ to generate instances based on high-level principles is shown in Table \ref{tab:prompt-generate}

\subsection{Human Evaluation}\label{sec:app:human}
We engaged three human annotators to label the data, following the same instruction as presented in \cite{chen2024alpagasus} (see Table \ref{tab:prompt-human}). Each annotator was required to choose the better response from the two options provided for each instruction. The response receiving the majority of votes was selected as the final result for each instruction.

\subsection{Analysis of Instances Length}

\begin{figure}[t]
        \centering
	\includegraphics[width=1 \linewidth]{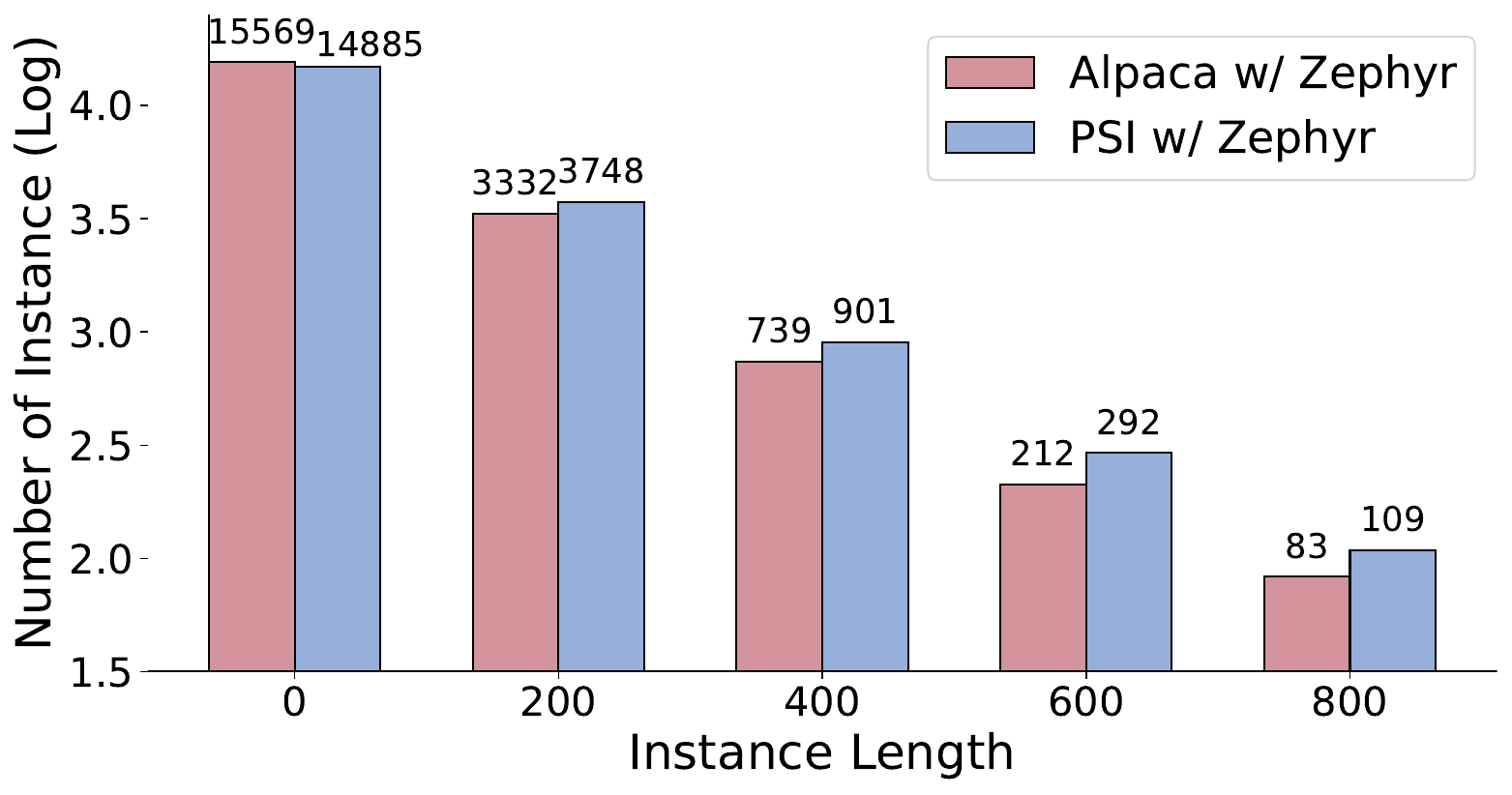}
        \caption{The instance length comparison between Alpaca and \model.}
        \label{fig:length-range}
\end{figure}

Some research works~\cite{longismore2024} have found that the length of the instruction tuning instances affects the performance.
Thus, in this section, we compare the sample length distribution of our \model and \texttt{Alpaca w/ Zephyr}.
As shown in Figure~\ref{fig:length-range}, we can find that the samples generated by our \model are longer than those generated by \texttt{Alpaca}.
It indicates the effectiveness of our proposed \second, which employs the task principles to guide the LLM to generate a high-quality dataset.

\subsection{Carbon Emission Calculation}~\label{app:carbon}

Based on previous research~\cite{patterson2021carbon,strubell-etal-2019-carbon-energy,dodge2022measuring-carbon}, we estimate the carbon emissions by calculating the total power required for generating and then multiplying it by the carbon emission intensity of the power grid used. 
It can be described using the following formula:

\begin{align*}
\small 
\text{Carbon Emissions}  &= \small \text{EC (kWh)} \times \small \text{CI (kgCO}_{2\text{e}}/\text{kWh)}\\
\end{align*}
where the EC is the electricity consumption, refers to the total amount of electrical energy used in the generating process, and CI represents the carbon intensity.
For a fair comparison, we use 0.24 $\text{kCO}_{2\text{e}}/\text{KWh}$, the carbon intensity in the Microsoft Azure US West region, for all following calculations.

Although reporting these operational emissions is standard practice, it overlooks other sources of emissions, such as those from the manufacturing, transportation, and disposal of hardware and data center infrastructure, lifetime operational emissions from usage, rebound effects, and other environmental impacts like water consumption and mining. Therefore, our estimates should be considered lower bounds.

For \texttt{Alpaca-GPT4}, due to the lack of detailed inference information disclosed by OpenAI, we can only make a preliminary estimation. We consider that ChatGPT require an estimated average electraicity consumption of 2.9 Wh per request based on \cite{DEVRIES20232191} . The average response generates 8 instances per request, so generating the 20k data we use requires 2.5k requests.
The carbon emission is :
\begin{equation*}
2.5k \times 2.9 \text{Wh} \times 0.24 \,  = 1.74 \, \text{kgCO}_{2\text{e}}
\end{equation*}

For \texttt{AlpaGasus}, it must first generate 52k instances, and then score all the instances using ChatGPT 52k times. Its carbon emission can be calculated as follows:
\begin{equation*}
(\frac{52}{8} + 52) \times 2.9  \times 0.24  = 40.72 \, \text{kgCO}_{2\text{e}}
\end{equation*}

For \texttt{WizardLM}, the author obtained the 250k instances with requesting ChatGPT 624k times in \cite{wizardlm2024xu}. We only use 20k of them, the carbon emission can be calculated as follows:
\begin{equation*}
624 \times \frac{20}{250} \times 2.9 \times 0.24  = 34.74 \, \text{kgCO}_{2\text{e}}
\end{equation*}

For our \model, we requested ChatGPT 10 times to reflect and summarize principles and ran a single A800 GPU with a power of 250W for 8 hours to generate instances.
The carbon emission is :
\begin{equation*}
\frac{10\times2.9+250\times8}{1000} \times 0.24 \,  = 0.49 \, \text{kgCO}_{2\text{e}}
\end{equation*}

\subsection{Learned Principles}\label{sec:principles}
In this section, we show the high-level principles learned by \model in Table \ref{tab:zephyr_p} and Table \ref{tab:llama3_p}.

\subsection{Compared with Principles Written Manually} ~\label{app:human_written}

To demonstrate that our method generates high-quality principles, we compared them with those written by humans. Specifically, we take three people four hours to manually create principles on a dataset of 1,000 instances. From the result in Table~\ref{tab:human_written}, we can observe that the principles generated by our multi-level principle generation framework, without filtering, have achieved a quality comparable to those created manually.

For tasks where human-written principles showed a decline in performance compared to those generated by PSI, we believe this reflects the inherent difficulty in manually crafting 1,000 instance-specific principles. The diversity and redundancy within the data make it challenging for human annotators to conduct a comprehensive analysis of all instances, thus limiting their ability to identify universally applicable improvements within the dataset. This further validates the effectiveness of our proposed multi-level principle generation framework, which can extract more generalizable improvement strategies in complex data environments.

\input{table/human_written}


\subsection{Implementation Details} ~\label{app:exp}

\paragraph{Training details}\label{sec:details:training} During the training stage, we follow the practices in instruction tuning in \texttt{Alpaca} \cite{alpaca}. We employed the  AdamW optimizer to finetune the model $\mathcal{M}_t$ for 3 epochs. The initial learning rate is set to $2 \times 10^{-5}$, with a  warm-up ratio of 0.03. The per GPU batch size is set to 8, resulting in a total batch size of 32, as we use 4 A800 GPUs for training. 

\paragraph{Cluster model}\label{sec:details:cluster} For clustering, we employ \texttt{all-mpnet-base-v2} in HuggingFace as the embedding model to encode the low-level principles $\mathcal{P}^L$. We then utilize the clustering algorithm described in~\cite{sarthi2024raptor}, which applies soft clustering without requiring a pre-defined number of clusters to organize these principles.

\begin{figure}[t]
        \centering
	\includegraphics[width=1\linewidth]{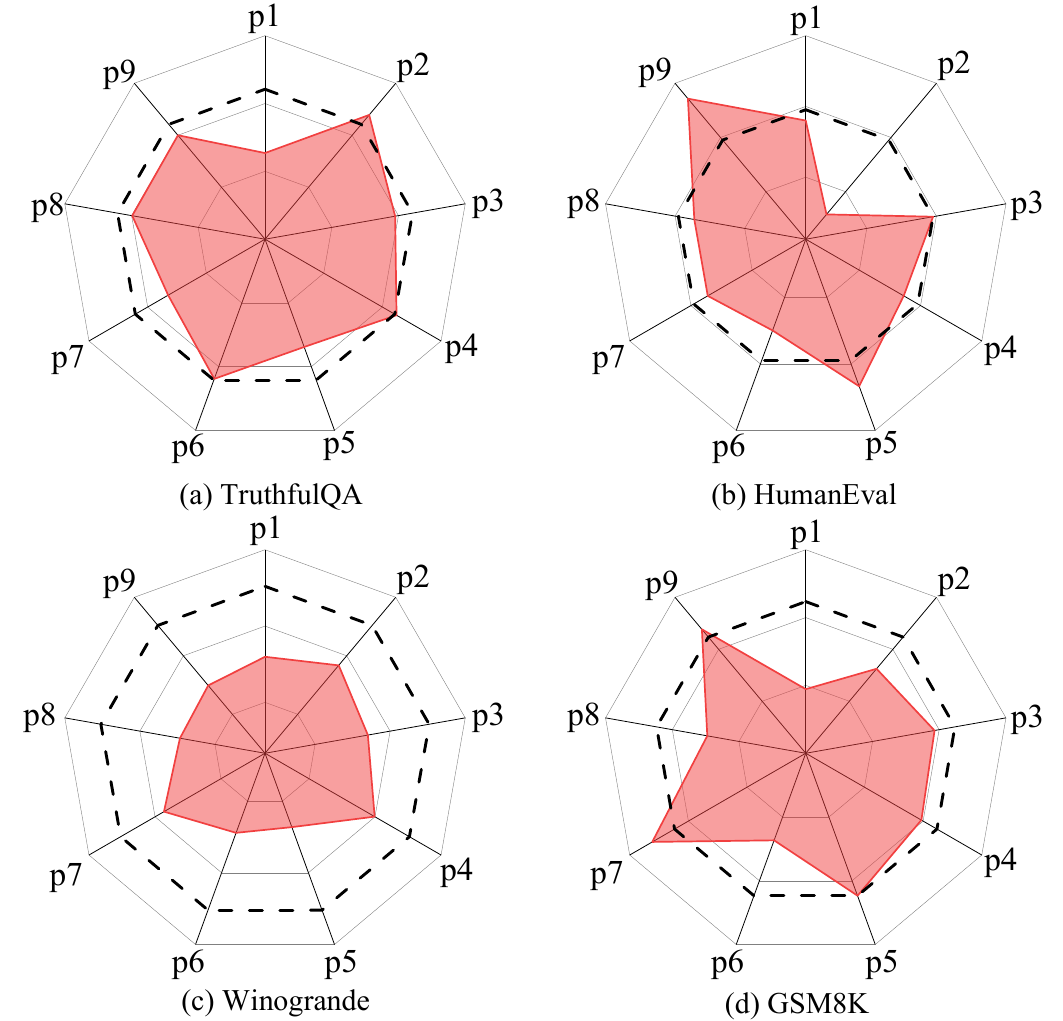}
        \caption{Effectiveness evaluation of each principle. Each vertex of the radar chart represents the performance of the model trained after removing the principle from the set. The dashed circle indicates the performance of the model using the whole set. }
 \label{fig:effect-principle-fulu}
\end{figure}

\subsection{Additional Experimental Results} ~\label{app:additional_exp}
In section~\ref{sec:exp:radar}, we only provide the results on two datasets after removing each principle. To obtain clearer and more comprehensive results, we provide the outcomes for four additional datasets in Figure~\ref{fig:effect-principle-fulu}. From the figure, we can see that $\mathcal{P}^H_1$ still plays an important role in the results of TruthfulQA and GSM8K. 

Through Figure 3 and Figure 7, we can analyze the relationship between principle selection and data representation when the $\mathcal{M}_g$ is Zephyr-7B. We can observe that Principle 1 (P1) is quite important for almost all tasks. P2 is helpful for generating data that enhances the model's reasoning capabilities in math and coding. Additionally, P3 aids in knowledge and commonsense reasoning. The remaining principles are mainly reflected in their significant help in a particular domain. For example, P8 can generate data that enhances mathematical abilities, while P4 is primarily helpful for commonsense reasoning.

\input{table/prompt}

\input{table/principle}

%% file: table/human_written.tex
\begin{table}[!t]
\centering
\small
    \setlength\tabcolsep{3pt}

\begin{tabular}{@{}l cc  cc  @{}}

\toprule

\textbf{Method}

& MMLU & GSM8K & HumanEval & GPQA
\\
\midrule

\textit{ $\mathcal{M}_t$: Gemma-2B}
\\

~+ \model  
& 41.7 &  19.0 & 36.6 & 26.3
\\

~+ Manual Writing
& 41.5 &  17.7 & 35.4 & 28.3
\\
\hdashline

\textit{ $\mathcal{M}_t$: Qwen-1.8B}
\\

~+ \model  
& 46.9 &  31.0 & 29.3 & 28.8
\\

~+ Manual Writing
& 45.6 &  29.2 & 31.7 & 26.7

\\
\bottomrule
\end{tabular}
\caption{Comparative analysis of principles in PSI and manual writing}
\label{tab:human_written}
\end{table}

%% file: table/prompt.tex
\begin{table*}[htbp]
\small \centering
\setlength\tabcolsep{4pt}
\begin{tabular}{@{}p{16cm}@{}}
\toprule
\textbf{Prompt to generate low-level principles }
\\
\hline

\\
The data instructions: \{INSTRUCTIONS\}.\\
\\
You are an AI assistant. Conduct a thorough analysis of the given instructions output pairs. Identify the points that can be improved.  Hallucinations, empty input and output, tasks that the language model cannot complete, etc. are all considered bad instructions.    Provide clear insights, principles, or guidelines that can be derived from this analysis to improve future instructions and outputs.    We are not focused on this one data point, but rather on the general principle.
 \\
 \\
Your output should follow the following format.\\
\\Reasoning: <discuss how the instruction generator could be improved>\\
\\
Insights: <what principle should be looked at carefully to improve the instructions outputs quality in the future, given in points>\\

\bottomrule
\end{tabular}
\caption{The prompt to generate low-level principles.}
\label{tab:prompt-low}
\end{table*}

\begin{table*}[htbp]
\small \centering
\setlength\tabcolsep{4pt}
\begin{tabular}{@{}p{16cm}@{}}
\toprule
\textbf{Prompt to generate high-level principles}
\\
\hline
\\
Low-level principles: \{low\_level\_principles\} \\
\\
Create a high level and  insightful principle to improve future responses based on the principles above.  Focus on capturing the essence of the feedback while eliminating redundancies.   Leave specific details in place.
 \\ 
\\
Principle:
\\
\bottomrule
\end{tabular}
\caption{The prompt to generate high-level principles.}
\label{tab:prompt-high}
\end{table*}

\begin{table*}[htbp]
\small \centering
\setlength\tabcolsep{4pt}
\begin{tabular}{@{}p{16cm}@{}}
\toprule
\textbf{Prompt to generate instances}
\\

\hline

\\
You are asked to come up with a set of 20 diverse task instructions. These task instructions will be given to a GPT model and we will evaluate the GPT model for completing the instructions.\\

Here are the requirements:\\
1. Try not to repeat the verb for each instruction to maximize diversity.\\
2. The language used for the instruction also should be diverse. For example, you should combine questions with imperative instrucitons.\\
3. The type of instructions should be diverse. The list should include diverse types of tasks like open-ended generation, classification, editing, etc.\\
2. A GPT language model should be able to complete the instruction. For example, do not ask the assistant to create any visual or audio output. For another example, do not ask the assistant to wake you up at 5pm or set a reminder because it cannot perform any action.\\
3. The instructions should be in English.\\
4. The instructions should be 1 to 2 sentences long. Either an imperative sentence or a question is permitted.\\
5. You should generate an appropriate input to the instruction. The input field should contain a specific example provided for the instruction. It should involve realistic data and should not contain simple placeholders. The input should provide substantial content to make the instruction challenging but should ideally not exceed 100 words.\\
6. Not all instructions require input. For example, when a instruction asks about some general information, "what is the highest peak in the world", it is not necssary to provide a specific context. In this case, we simply put "<noinput>" in the input field.\\
7. The output should be an appropriate response to the instruction and the input. Make sure the output is less than 100 words.\\
\\
The following insights and guidelines may improve responses:\\
\{PRINCIPLES\}\\
\\
List of 20 tasks:
\\

\bottomrule
\end{tabular}
\caption{The prompt to generate instances.}   
\label{tab:prompt-generate}
\end{table*}

\begin{table*}[htbp]
\small \centering
\setlength\tabcolsep{4pt}
\begin{tabular}{@{}p{16cm}@{}}
\toprule
\textbf{Evaluation criterion for human annotators }
\\
\hline
\\
You’ll be presented with a series of questions. For each question, two answers will be provided. Your task is to read both answers carefully and decide which one you believe is better. When judging, consider: \\
\\
\textbf{Relevance:} Does the answer directly address the question? \\
\textbf{Completeness:} Is the answer comprehensive? \\
\textbf{Coherence:} Is the answer logically structured and easy to understand? \\
\textbf{Accuracy:} Is the information provided in the answer correct? \\
\\
\textbf{Question:} <QUESTION>  \\
\textbf{Answer A:}  <ANSWER A>  \quad \textbf{Answer B:} <ANSWER B> \\
\\
\textbf{Comparing these two answers, which answer is better? }\\
1. Answer A is significantly better.  \\
2. Answer B is significantly better.  \\
3. Neither is significantly better. \\
\\
\bottomrule
\end{tabular}
\caption{The detailed evaluation criterion for human annotators.}
\label{tab:prompt-human}
\end{table*}

%% file: table/principle.tex
\begin{table*}[htbp]
\small \centering
\setlength\tabcolsep{4pt}
\begin{tabular}{@{}p{16cm}@{}}
\toprule
\textbf{Learned high-level principles for Zephyr-7B-beta by PSI}
\\
\hline
\\
1. Ensure clear and concise communication by aligning the output with the prompt, maintaining proper formatting and grammar, addressing the task accurately and comprehensively, and organizing the output logically for better readability.\\
2. Consider the language model's limitations, provide clear and specific instructions, and anticipate potential constraints to ensure accurate and feasible responses.\\
3. Ensure accurate and concise paraphrases or summaries by including relevant information and guiding the model to provide simpler and more straightforward responses.\\
4. Ensure clear and relevant inputs to guide the AI in generating accurate and meaningful outputs.\\
5. Ensure clear and concise instructions that provide complete and specific information, avoiding ambiguity and open-endedness, while encouraging creativity and providing examples or templates when necessary.\\
6. Provide clear and detailed instructions that address the specific context and requirements of the task, including specific guidelines, prompts, and background information, to guide the generation of relevant and inclusive outputs.\\
7. Ensure comprehensive and accurate responses by understanding and incorporating user preferences, adhering to given restrictions, and providing detailed feedback, while avoiding fictional or irrelevant information.\\
8. Ensure meaningful and complete inputs and outputs to enhance the AI's understanding and delivery of relevant information.\\
9. Provide clear and concise instructions that guide the model to generate accurate, comprehensive, and effective outputs, while avoiding unsupported assumptions or hallucinations.\\

\\
\bottomrule
\end{tabular}
\caption{The high-level principles for Zephyr-7B-beta.}
\label{tab:zephyr_p}
\end{table*}

\begin{table*}[htbp]
\small \centering
\setlength\tabcolsep{4pt}
\begin{tabular}{@{}p{16cm}@{}}
\toprule
\textbf{Learned high-level principles for Llama3-8B-Instruct by PSI}
\\
\hline
\\
1. Encourage comprehensive and critical analysis by providing clear instructions that include all relevant details, specify evaluation criteria, and request justification or reasoning, while also ensuring outputs include explanations or examples, final results when applicable, and distinct explanations of differences or similarities.\\
2. Provide clear and specific instructions: Clear and specific instructions, with relevant context and desired outcomes, help the AI assistant understand the task accurately and generate more accurate and meaningful responses. Avoid ambiguity and subjective interpretations by providing objective guidelines and criteria.\\
3. Continuous improvement through user feedback and iterative refinement enhances the quality, accuracy, and relevance of AI assistant responses, ensuring a better user experience and reliable assistance.\\
4. Continuous improvement: By continuously testing, iterating, and fact-checking instructions and outputs, while ensuring alignment between input and output, future responses can be improved to provide accurate, useful, and personalized information. Incorporating user feedback and considering creativity in the generation process will lead to better instruction outputs over time.\\
5. Promote creativity and engagement by providing clear guidelines and constraints, specifying format and content expectations, and offering specific prompts or guidelines for tasks that require creative writing.\\
6. Provide clear and comprehensive context: Instructions should include relevant context, examples, and specific details to guide the AI assistant in generating accurate and informative responses.\\
7. Align instructions with the language model's capabilities and limitations to ensure accurate and meaningful responses.\\
8. Ensure accurate and relevant responses by providing complete and specific instructions, avoiding hallucinations or unsupported claims, and maintaining consistency and coherence between inputs and outputs.\\
9. Clear and specific instructions: Instructions should provide clear and specific details to guide the model's response, including the desired outcome, required input format, expected output format, and any specific elements or explanations needed. Ambiguity should be avoided, and guidelines for multiple definitions or categorizations should be clarified.\\

\\
\bottomrule
\end{tabular}
\caption{The high-level principles for Llama3-8B-Instruct.}
\label{tab:llama3_p}
\end{table*}